\documentclass[10pt,twocolumn,letterpaper]{article}

\usepackage{cvpr} 
\usepackage{times}
\usepackage{epsfig}
\usepackage{graphicx}
\usepackage{amsmath}
\usepackage{amssymb}
\usepackage{balance}

\usepackage[pagebackref=true,breaklinks=true,letterpaper=true,colorlinks,bookmarks=false]{hyperref}

\usepackage{times}
\usepackage{epsfig}
\usepackage{amssymb}
\usepackage{booktabs}
\usepackage{array}
\usepackage{amsmath}
\usepackage{txfonts}
\usepackage{mathdots}
\usepackage{balance}
\usepackage{multirow}
\usepackage{graphicx}
\usepackage[table]{xcolor}
\graphicspath{ {imgs/} }
\usepackage{algorithm}
\usepackage{algpseudocode} 
\usepackage{pifont}   
\usepackage{subcaption}
\usepackage{multirow}
\usepackage{caption}
\usepackage{capt-of,etoolbox}
\usepackage{algorithm}
\usepackage{algpseudocode}
\usepackage{placeins}
\usepackage{mwe}
\usepackage{lipsum}
\usepackage{cuted}  
\usepackage{comment}

\newcommand{\cmark}{\ding{51}} 
\newcommand{\xmark}{\ding{55}} 

\begin{document}

\title{DOOMGAN: High-Fidelity Dynamic Identity Obfuscation Ocular Generative Morphing}

\author{Bharath Krishnamurthy and Ajita Rattani \\
University of North Texas\\
Denton, Texas, USA\\
{\tt\small BharathKrishnamurthy@my.unt.edu, Ajita.Rattani@unt.edu}}

\maketitle
\begin{abstract}
Ocular biometrics in the visible spectrum have emerged as a prominent modality due to their high accuracy, resistance to spoofing, and non-invasive nature. 
However, morphing attacks, synthetic biometric traits created by blending features from multiple individuals, threaten biometric system integrity. While extensively studied for near-infrared iris and face biometrics, morphing in visible-spectrum ocular data remains underexplored. Simulating such attacks demands advanced generation models that handle uncontrolled conditions while preserving detailed ocular features like iris boundaries and periocular textures. To address this gap, we introduce \textbf{DOOMGAN}, that encompasses landmark-driven encoding of visible ocular anatomy, attention-guided generation for realistic morph synthesis, and dynamic weighting of multi-faceted losses for optimized convergence. DOOMGAN achieves over 20\% higher attack success rates than baseline methods under stringent thresholds, along with 20\% better elliptical iris structure generation and 30\% improved gaze consistency. We also release the first comprehensive ocular morphing dataset to support further research in this domain. The code is available at~\href{https://github.com/vcbsl/DOOMGAN}{Vcbsl/DOOMGAN}.

\end{abstract}


\section{Introduction}
\begin{figure}[ht!]
  \centering
   \includegraphics[width=0.90\linewidth]{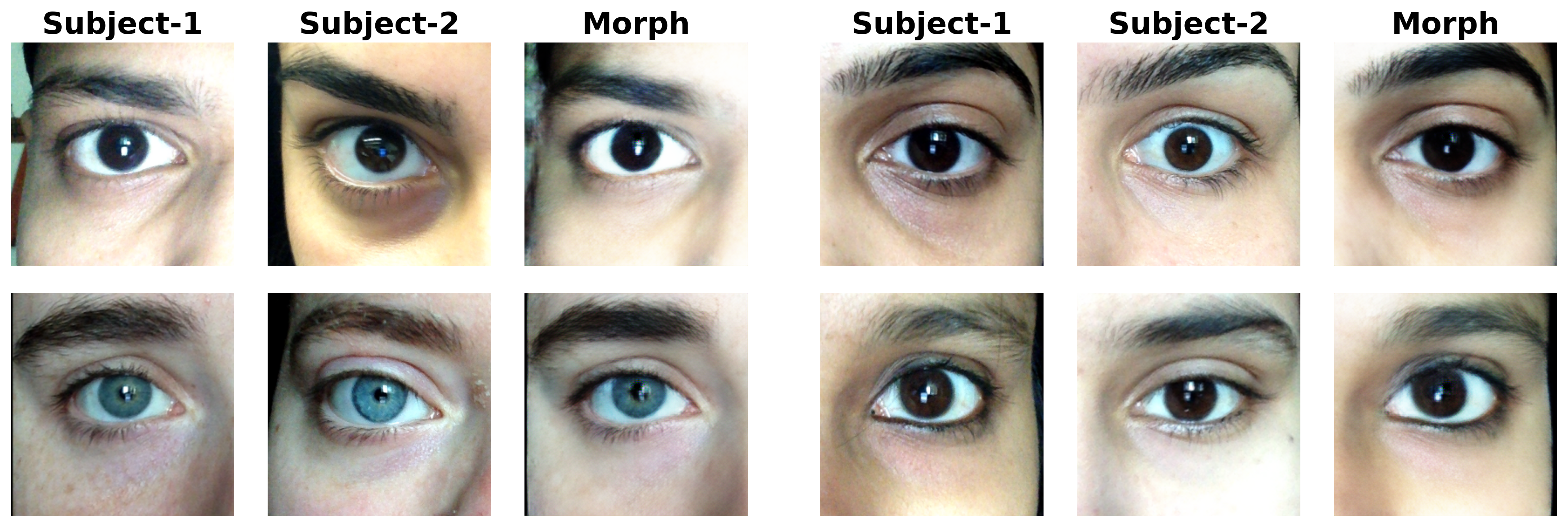}
   \captionof{figure}{Illustration of morphed ocular images generated by our proposed "DOOMGAN". A morphed image combines images of two different identities (subjects) such that the ensuing biometric template can match both contributing identities, posing a serious security threat to biometric systems.}
    \label{fig:teaser}
\end{figure}

Ocular biometrics has received significant attention from academia and industry alike, for its identity resolving power and security~\cite{rattani2017ocular, jain2024multibiometrics, raja2022towards, krishnamurthy2025abrobocular}. Ocular biometrics in the visible spectrum include vascular patterns seen on the white of the eye, eyebrows, and the periocular region (mostly skin texture and wrinkles) encompassing the eye. Further, visible light ocular modality can be acquired using almost any imaging device, ranging from high-end dSLR to ubiquitous mobile device cameras, which is particularly advantageous. 

Despite the advantages associated with ocular biometrics in the visible spectrum, the low quality of images captured in unconstrained conditions (i.e., uncontrolled environment) remains challenging due to factors such as eyelashes, gaze deviation, varying illumination, and specular reflection. Further, visible light ocular images may demonstrate low contrast between the iris and pupil boundary, especially for dark pigmented irides. This is because photic accumulation in the visible spectrum is so low that texture may be lost in sensor noise. Consequently, lack of texture, incorrect localization, and errors in segmentation leads to lower visible light ocular recognition accuracy. Accordingly, a number of studies have been proposed for generalized and robust feature extraction techniques for visible light ocular-based user recognition
~\cite{reddy2018ocularnet, rattani2018convolutional, reddy2019robust, reddy2020generalizable}.

Unsupervised biometric enrollment has gained significant traction due to the increasing demand for remote authentication services, accelerated by global digitalization and the need for contactless solutions in various sectors, including banking, healthcare, and government services~\cite{jain2024multibiometrics, raja2022towards}. However, unsupervised enrollment poses a severe security risk, where the enrolling subject submits the \emph{morphed} image to the system referred to as a \emph{Morphing Attack}. A morphed image combines images of two different individuals (subjects), and the ensuing biometric template can potentially match both contributing identities. Thus, undermining the \textbf{trustworthiness of the biometric technology}. 

Morphing attacks have been studied for face~\cite{chang2005evaluation, meden2021privacy, rattani2018survey} and NIR-iris biometrics~\cite{wang2003combining, bowyer2008image, nguyen2024deep}. The proposed facial morph generation methods are particularly effective when the pair of images are acquired in controlled acquisition scenarios involving visually similar contributing subjects~\cite{raghavendra2017face, damer2018morgan, venkatesh2020can, venkatesh2021face,pn2024morcode}. Normalized irises are morphed using traditional computer vision techniques~\cite{rathgeb2017feasibility, Erdogan, sharma2021image} at the NIR spectrum. However, the implications of morphing attacks on VIS ocular biometrics remain largely \textbf{unexamined}.


\noindent\textbf{Challenges.} Ocular morphing in the visible spectrum presents several critical challenges:
\begin{itemize}
\item \textbf{Operation in Uncontrolled Acquisition Environments:} Ocular images in the visible spectrum are often captured under unconstrained conditions~\cite{proencca2016unconstrained, wang2020towards} complicating the reliable generation of ocular morphs in the visible spectrum~\cite{rattani2017ocular}.
\item \textbf{Preservation of Fine-Grained Details:} High-quality VIS ocular morphs demand accurate synthesis of fine-grained features, such as detailed skin texture and the distinctive elliptical shape of the iris within the eyelids~\cite{guo2022eyes}. Maintaining the integrity of these fine-grained details during morph generation remains a key technical challenge~\cite{yadav2019synthesizing, yadav2024synthesizing}.
\item \textbf{Generalization Across Identities:} Unlike face morphing, where visually similar subjects from the same demographic group are often used, ocular images lack clear identity and demographic cues, demanding generalizability across a diverse range of subjects and demographics~\cite{venkatesh2020can, alonso2024periocular}. 
\item \textbf{Enhanced Risk of Mode Collapse:} GAN-based models are particularly prone to mode collapse when faced with the structural complexity and fine-grained details inherent in ocular images~\cite{thanh2020catastrophic, srivastava2017veegan, gulrajani2017improved}. This limits the variability and authenticity of the generated morphs, posing a substantial barrier to the robustness of the model.
\end{itemize}

\noindent\textbf{Contributions.} We propose DOOMGAN, a novel generative adversarial network architecture designed to tackle the significant challenges of generating high-fidelity, identity-obscuring morphs for visible-spectrum (VIS) ocular biometrics, particularly in uncontrolled environments, as illustrated in Figure~\ref{fig:teaser}. DOOMGAN features a landmark-guided, attention-based Generator, enabled by a novel Ocular Landmark Generator, and incorporates a spectrally normalized Encoder utilizing Gaussian landmark heatmaps. It also employs a patch-wise Discriminator to enforce local realism. Training is guided by a multi-faceted loss function that integrates structural similarity, perceptual loss, and identity difference, with their contributions adaptively balanced using a novel dynamic weighting mechanism. The contributions of our work are summarized as follows:

\begin{enumerate}
\item \textbf{A Novel Ocular Landmark Generator:} We introduced a novel ocular landmark generator that is inspired from and surpasses existing Mediapipe-Iris, an ocular landmark generator from Google~\cite{ablavatski2020real}  in precision and versatility. This is crucial for the accurate morphing of anatomical structures in ocular regions, while also ensuring robustness against gaze deviations and off-angle iris regions. 

\item \textbf{A Specialized Ocular Morphing Model:} We propose DOOMGAN, a novel GAN-based framework designed specifically for synthesizing high-quality ocular morphs. Our model incorporates several key innovations: landmark-driven encoding to capture the complex anatomical structure of the eye, attention-guided generation, and dynamic weighting for optimized convergence~\cite{zhang2019self}. 

\item \textbf{A Novel Dynamic Weighting based Training Scheme for Multi-faceted Loss Function:} We propose an innovative dynamic weighting process to balance the contributions of each loss function, systematically adjusted to facilitate simultaneous convergence of all loss functions during training.

\item \textbf{Novel Performance Metrics and Thorough Vulnerability Assessment:}
We proposed novel ocular-specific metrics, Iris Irregularity (IR) and Gaze Direction (Gaze) consistency, to assess DOOMGAN in capturing fine-grained features measured by elliptical iris boundaries and consistent gaze direction. We evaluated the effectiveness of DOOMGAN by testing its vulnerability against existing ocular recognition algorithms using four benchmark datasets. Further, the efficacy of the existing Morph Attack Detection~(MAD), based on texture descriptors and deep-learning-based models~\cite{scherhag2020deep, ferrara2022morphing}, in detecting the generated VIS ocular morphs is also evaluated. 

\item \textbf{Contributing First Ocular Morph Datasets and Tools}: We contribute extensive visible light \textbf{morphed ocular image datasets} with over $1,00,000$ high-quality morphed images synthesized from benchmark datasets. Thus, facilitating further research into detecting and defending against morphing attacks in VIS ocular biometrics.
\end{enumerate}

\section{Related Work}
\label{related}
\noindent\textbf{Face:} 
Morphing attack generation has been primarily investigated for facial biometrics. Mostly, GAN-based models using interpolation of the facial embeddings have been used for facial morph generation. The advances follow MorGAN~\cite{damer2018morgan}, a GAN-based model generating low-resolution morphs to StyleGAN (higher resolution)~\cite{venkatesh2020can}, VAE-GAN~\cite{larsen2016autoencoding, jiang2021geometrically}, employing encoder structure to obtain latent representation, MIPGAN~\cite{zhang2021mipgan}, based on ResNet encoder and utilizing identity losses, and the latest being MorCode~\cite{pn2024morcode}, leverages VQGAN with learned discrete codebooks and spherical interpolation to blend latent representations. However, these generative models necessitate visual similarity of the facial images of the contributing subjects captured in a controlled environment. Most of these models require post-processing of the generated morphed output for artifacts removal. Apart from these, handful of studies also utilized diffusion model for facial morph generation~\cite{damer2023mordiff, grimmer2024ladimo}.

\textbf{NIR-Iris:} In the case of the iris modality, existing approaches have primarily employed \textit{traditional computer vision} techniques.~\cite{rathgeb2017feasibility} proposed morphing at the feature level where iris codes (i.e., binary representation of the unique features extracted from an iris image after it has been normalized to a fixed size and shape), are morphed using stability-based bit substitution. ~\cite{Erdogan} proposed an iris morphing approach using normalized iris images (i.e., unwrapped iris to a fixed-size rectangular entity) by selecting pixels from two normalized iris images based on their intensity and phase profiles. ~\cite{sharma2021image} proposes iris morphing in the NIR spectrum using the landmark-based Delaunay triangulation technique on the unnormalized images. Readers are referred to Section $1$ in the supplementary material for further details on NIR-Iris Morphing.

\textbf{MAD:} In an attempt to detect morphing attacks, recent studies have proposed several Morph Attack Detection~(MAD) algorithms that often utilize texture descriptors, such as Local Binary Patterns~(LBP)~\cite{pietikainen2010local} and Histogram of Oriented Gradients~(HOG)~\cite{dalal2005histograms,ahonen2008recognition} along with traditional machine learning classifiers, and deep learning-based convolutional neural networks~\cite{venkatesh2021face} and vision transformers~\cite{zhang2024generalized}.

\section{Method}
\subsection{Architectural Details}

\begin{figure}[ht!]
   \centering
   \includegraphics[width=0.85\linewidth]{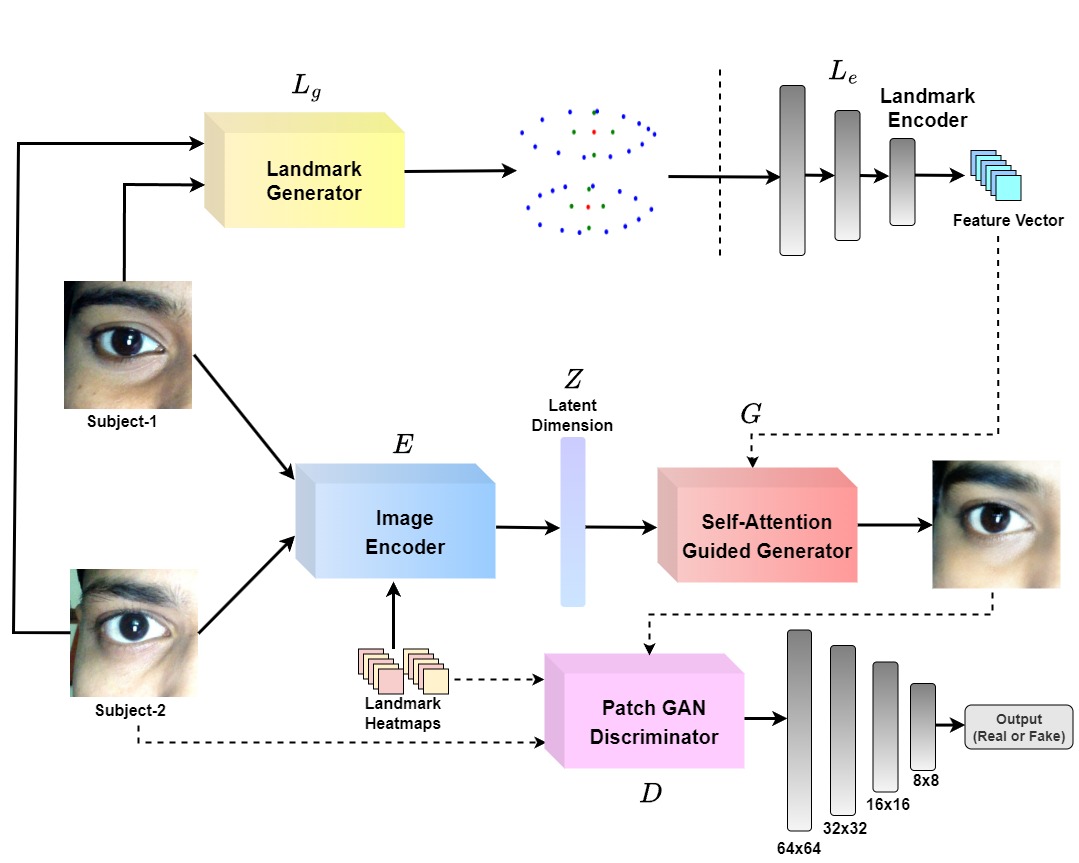}
   \caption{The overall architecture of the proposed DOOMGAN for visible spectrum ocular morph generation.}
   \label{fig:Morph_Preview}
\end{figure}

Figure~\ref{fig:Morph_Preview} shows an overview of the architecture of our proposed model to generate morphed ocular images. The architecture comprises five essential components, namely, \textbf{Landmark Generator} (\(L_g\)), \textbf{Landmark Encoder} (\(L_e\)), \textbf{Image Encoder using Landmark Heatmaps} (\(E\)), \textbf{Attention-Based Landmark Guided Generator} (\(G\)), and \textbf{Spectral Normalization Powered Discriminator} (\(D\)).

For each batch of paired ocular image data from the contributing subjects, we generate the corresponding landmark data using $L_g$. The extracted landmarks are used to create landmark heatmaps to assist the $E$ and the $D$. The landmarks are further passed through the landmark encoder $L_e$ to extract the feature vector encoding their spatial coordinates, to guide the $G$ to create plausible ocular morphs. The $E$ creates the corresponding latent dimensional space for the ocular image data, assisted by landmark heatmaps. The $E$, $G$, and $D$ are simultaneously trained to generate high-fidelity ocular morphs, capable of bypassing biometric systems at numerous latent disentanglement points. Subsequently, we discuss each of these components in detail:

\vspace{1em}
\vspace{-2em}
\subsubsection{Landmark Generator (\(L_g\))}
Our proposed Ocular Landmark Generator $L_g$ represents a significant advancement in the field of ocular biometrics, offering several key improvements over the latest Mediapipe IRIS~\cite{ablavatski2020real, chavarro2024blink}. Let $I \in \mathbb{R}^{H \times W \times 3}$ represent an input ocular image, where $H$ and $W$ are the height and width of the image, respectively. The $L_g$ can be represented as a function $f_\theta: \mathbb{R}^{H \times W \times 3} \rightarrow \mathbb{R}^{2N}$, where $\theta$ represents the learnable parameters of the neural network, and $N$ is the number of landmarks. The generator output is a vector of 2D coordinates, $\mathbf{L} = {\{(x_1, y_1), \ldots, (x_N, y_N)\}}$, representing the predicted landmarks. The function $f$ is a CNN that consists of two convolutional blocks for feature extraction, followed by two fully connected layers and the output layer of $66$ neurons for the prediction of landmark coordinates. Each convolutional block for the $L_g$ contains a convolutional layer, followed by ReLU activation, and a max-pooling layer. $L_g$ is trained using the mean squared error (MSE) loss between the predicted and the ground truth landmarks. The MSE of $L_g$ was $0.54$ when evaluated on the VISOB~\cite{rattani2016icip} test set. Readers are referred to Section $2$ in the supplementary material for further details on our proposed $L_g$.
\vspace{-1em}
\subsubsection{Landmark Encoder (\(L_e\))}

The Landmark Encoder \(L_e\) processes the generated landmarks $(l)$ to extract meaningful feature representation $f_l$ of dimension $D_l$. It employs linear layers with Leaky ReLU activation to capture the spatial relationships of the landmarks. By transforming the landmark coordinates into the $f_l$, \(L_e\) enhances the generator's ability to incorporate both geometric and appearance features of the ocular images.
\begin{equation}
    f_l = L_e(l), \quad f_l \in \mathbb{R}^{D_l}
\end{equation}
Moreover, we define \emph{landmark heatmaps} ($h_l$) as Gaussian heatmaps ($H(l)$) centered on the location of the landmark coordinates. These heatmaps, obtained from $L_g$, are essential for integrating spatial landmark information ($l$). A total of $N_h$ heatmaps ($19$ heatmaps were chosen, determined by the total number of landmarks encompassing the regions of the eye and iris, in order to provide sufficient anatomical guidance to Generator) are generated and concatenated with the input images (of size $H \times W$) in $E$ and $D$ to provide additional context on the regions of importance, such as the iris and eyebrow regions.
\begin{equation}
    h_l = H(l), \quad h_l \in \mathbb{R}^{N_h \times H \times W}
\end{equation}
\vspace{-2em}
\noindent\subsubsection{Image Encoder (\(E\))}
The Image Encoder \(E\) is tasked with transforming the input ocular images \(x\), augmented with corresponding landmark heatmaps \(h_l\), into a compact latent representation \(z\) of dimension $D_z$. By incorporating the landmark heatmaps directly via concatenation, \(E\) is designed to capture both the visual appearance features from the image and the explicit spatial configuration defined by the landmarks, producing an enriched latent space highly conducive for generating realistic morphs. 
\(E\) uses a combination of specialized blocks: \textbf{Encoder blocks}, each comprising a \textit{spectral normalized convolution}, \textit{instance normalization}, and \textit{LeakyReLU activation}, progressively downsizing the spatial dimensions while increasing feature channels. \textit{Residual blocks} employing \textit{skip connections}, \textit{convolution}, \textit{instance normalization}, and \textit{ReLU} are utilized to learn optimized residual mappings and mitigate potential vanishing gradients. Notably, spectral normalization is omitted in the first encoder block to allow for more flexible initial feature learning, while subsequent blocks leverage it for enhanced training stability. Furthermore, a strategically placed \textit{self-attention module} captures long-range dependencies within the feature maps, crucial for understanding complex ocular structures. This entire process culminates in the latent representation \(z\), vital for guiding the generator.
\begin{equation}
    z = E(x, h_l), \quad z \in \mathbb{R}^{D_z}
\end{equation}
\vspace{-2em}
\noindent \subsubsection{Attention-Based Landmark Guided Generator}
The Generator \(G\) synthesizes the final morphed ocular image \(\hat{x_{morph}}\) by decoding the combined latent representation \(z\) from the \(E\), critically guided by the landmark features \(f_l\) provided by the Landmark Encoder \(L_e\). By integrating this landmark feature information ($f_l$), \(G\) is guided to generate high-quality morphs that blend the features of two identities. At its core, the $G$ employs an \textit{attention mechanism} to focus on important features guided by the landmarks, emphasizing key anatomical regions like the iris and eyebrows for realistic morphs. $G$ is guided to generate high-quality morphs ($\hat{x_{morph}}$) that blend the features of two identities through a series of \textbf{generator blocks}, each utilizing \textit{transposed convolution} for upsampling, \textit{instance normalization}, and \textit{ReLU activation}. \textit{Residual blocks} are integrated, particularly near the attention module, to stabilize the training process and facilitate robust information flow, ultimately enabling the generation of high-quality morphs that accurately blend the features and anatomical characteristics of the two source identities.
\begin{equation}
    \hat{x_{morph}} = G(z, f_l)
\end{equation}

\subsubsection{Spectral Normalization Powered Discriminator}
The Discriminator \(D\) plays a crucial role in adversarial training, distinguishing real images from generated ones. It incorporates spectral normalization to stabilize the training by controlling the Lipschitz constant of the network. \(D\) also utilizes gradient penalty and a PatchGAN-based architecture with multiple outputs to effectively evaluate local patches of the image. The $D$ produces multi-scale real ($y_i$) and fake ($\tilde y_i$) scores, providing rich feedback to the generator that encourages the synthesis of more realistic images.
\begin{equation}
\begin{aligned}
D_i(x,\,h_l) &= \{\,y_1,\,y_2,\,y_3,\,y_4\}, \\
D_i(\hat x_{\mathrm{morph}},\,h_l) &= \{\tilde y_1,\,\tilde y_2,\,\tilde y_3,\,\tilde y_4\},
\end{aligned}
\quad
y_i,\;\tilde y_i\;\in\;\mathbb{R}.
\end{equation}
The $D$ network comprises of the \textbf{discriminator blocks}, each incorporating a \textit{spectral normalized convolution} layer for stable training through 2D convolution with normalized weights. The \textit{first discriminator block} omits spectral normalization to ensure the capture of complete information from the input. The network outputs feature maps at progressively smaller scales—$64 \times 64$, $32 \times 32$, $16 \times 16$, and $8 \times 8$—providing a multi-scale representation that enhances the analysis of both local and global features.



\subsection{Loss Functions}
To enhance the training process, we employ a combination of loss functions with dynamic weight adjustment to achieve high-quality ocular morphing. Unlike prior studies~\cite{venkatesh2020can, zhang2021mipgan} that rely on constant weights $\lambda$ determined empirically, our approach introduces a dynamic weighting mechanism to balance loss contributions adaptively during training. This innovation reduces the need for manual intervention, accelerates convergence, and improves image quality. Below, we detail the role of each loss function, inspired from the studies~\cite{arjovsky2017wasserstein,wang2003multiscale,johnson2016perceptual,zhang2021mipgan,damer2018realistic}, and the dynamic weighting process integrated into our training framework.
\begin{enumerate}
    \item \textbf{Adversarial Loss (Base Loss)}: We employ a multi-scale discriminator approach:
        \begin{equation}
            \mathcal{L}_{\text{adv}} = \mathbb{E}_{x \sim p_{\text{data}}}[-D(x)] + \mathbb{E}_{z \sim p_z}[D(G(z))]
        \end{equation}
        This adversarial loss is based on the Wasserstein GAN formulation.

    \item \textbf{MS-SSIM Loss}: To ensure structural similarity across multiple scales:
        \begin{equation}
            \mathcal{L}_{\text{MS-SSIM}} = 1 - \text{MS-SSIM}(x, G(E(x)))
        \end{equation}

    \item \textbf{Perceptual Loss}: We further aim to enhance the visual quality by using perceptual loss with a pre-trained VGG network $\phi$ at various intermediate layers:
        \begin{equation}
            \mathcal{L}_{\text{perc}} = \sum_{i \in \text{loss\_layers}} \|\phi_i(x) - \phi_i(G(E(x)))\|_2^2
        \end{equation}

    \item \textbf{Reconstruction Loss}: To ensure accurate reconstruction:
        \begin{equation}
            \mathcal{L}_{\text{rec}} = \|x - G(E(x))\|_2^2
        \end{equation}
        The reconstruction loss is a standard L2 loss, commonly used in various deep-learning applications.

    \item \textbf{Identity Loss}: To maintain identity-related features, we utilize a pre-trained ocular recognition model based on the ResNet-50 backbone and the ArcFace loss function. The identity loss is computed using the cosine similarity of the embedding of the morphed image to the embeddings from the pair of images from the contributing subjects:
        \begin{equation}
            \mathcal{L}_{\text{Id}} = 1 - \frac{1}{2} \left( \frac{f(x_1) \cdot f(G(z))}{\| f(x_1) \| \| f(G(z)) \|} + \frac{f(x_2) \cdot f(G(z))}{\| f(x_2) \| \| f(G(z)) \|} \right)
        \end{equation}
        where $f$ is a pre-trained ocular recognition model, $x_1$ and $x_2$ are the two source identities, and $G(z)$ is the resulting morph.

    \item \textbf{Identity Difference Loss}: To balance identity preservation and solve the imbalance between different subjects, we incorporate this loss as follows:
        \begin{equation}
            \mathcal{L}_{\text{id\_diff}} = \frac{f(x_2) \cdot f(G(z))}{\|f(x_2)\| \| f(G(z)) \|} - \frac{f(x_1) \cdot f(G(z))}{\|f(x_1)\| \| f(G(z)) \|}
        \end{equation}
        This loss is inspired by the work on face morphing and identity preservation.
\end{enumerate}

The total loss of DOOMGAN is a weighted sum of these component losses:
\begin{equation}
    \mathcal{L}_{\text{total}} = \sum_{i} w_i \mathcal{L}_i
\end{equation}
where $w_i$ are dynamically adjusted weights for each loss component.

\begin{table*}[ht]
\small
\centering
\setlength{\tabcolsep}{2.5pt}
\renewcommand{\arraystretch}{0.9}
\begin{tabular}{l|c|c|c|c|ccc|ccc|ccc|ccc}
\hline
& & & & & \multicolumn{6}{c|}{\textbf{OVS-I}} & \multicolumn{6}{c}{\textbf{OVS-II}} \\
\hline
\textbf{Method} 
  & \textbf{SSIM} & \textbf{IR} & \textbf{Gaze} & \textbf{Inf.}
  & \multicolumn{3}{c|}{\underline{\textbf{MMPMR}}} 
  & \multicolumn{3}{c|}{\underline{\textbf{FMMPMR}}} 
  & \multicolumn{3}{c|}{\underline{\textbf{MMPMR}}} 
  & \multicolumn{3}{c}{\underline{\textbf{FMMPMR}}} \\
& & & &
  & \textbf{0.01\%} & \textbf{0.1\%} & \textbf{1\%} 
  & \textbf{0.01\%} & \textbf{0.1\%} & \textbf{1\%} 
  & \textbf{0.01\%} & \textbf{0.1\%} & \textbf{1\%} 
  & \textbf{0.01\%} & \textbf{0.1\%} & \textbf{1\%} \\
\hline
LM~\cite{sharma2021image}        
  & \textbf{0.72} & 0.7328 & 0.5916 & -
  & \textbf{92.50} & 94.52 & 99.76 
  & 15.32 & 44.76 & 79.58 
  & \textbf{96.09} & \textbf{98.79} & 99.97 
  & 7.54 & 34.17 & 89.16 \\
MorGAN~\cite{damer2018morgan}    
  & 0.46 & 0.5938 & 0.4716 & 2.86      
  & 3.92 & 33.60 & 65.43 
  & 0.01 & 2.31 & 34.64 
  & 6.47 & 18.98 & 94.53 
  & 0.02 & 9.87 & 72.89 \\
StyleGAN~\cite{venkatesh2020can}  
  & 0.47 & 0.6683 & 0.5080 & 30.57
  & 4.60 & 33.62 & 84.55 
  & 0.11 & 3.89 & 37.77 
  & 12.05 & 34.52 & 98.10 
  & 0.39 & 3.76 & 75.36 \\
VAE-GAN~\cite{larsen2016autoencoding}
  & 0.56 & 0.7099 & 0.6548 & 5.87
  & 26.38 & 73.14 & 91.57 
  & 0.23 & 9.28 & 52.76 
  & 34.78 & 71.71 & 99.98
  & 1.29 & 11.91 & 92.36 \\
MIPGAN~\cite{zhang2021mipgan}    
  & 0.46 & 0.7723 & 0.5753 & 2764
  & 28.22 & 72.73 & 97.61 
  & 2.39 & 23.18 & 71.33 
  & 35.17 & 64.07 & 99.25 
  & 3.80 & 15.48 & 87.83 \\
MorCode~\cite{pn2024morcode}      
  & 0.54 & 0.7826 & 0.7577 & 14.17
  & 32.42 & 77.48 & 98.46 
  & 3.84 & 28.21 & 76.85 
  & 37.97 & 82.90 & 99.79 
  & 3.98 & 27.59 & 88.44 \\
\textbf{DOOMGAN}                 
  & 0.67 & \textbf{0.9380} & \textbf{0.9775} & 7.34
  & 75.00 & \textbf{95.73} & \textbf{99.85} 
  & \textbf{30.95} & \textbf{70.57} & \textbf{96.33} 
  & 86.15 & 96.33 & \textbf{100} 
  & \textbf{38.71} & \textbf{75.20} & \textbf{93.01} \\
\hline
\end{tabular}
\caption{Comparative evaluation of DOOMGAN against landmark-based (LM) and GAN-based (MorGAN, StyleGAN, VAE-GAN, MIPGAN, and MorCode) morphing models on VISOB, assessing their impact on OVS-I and OVS-II at various operating thresholds. The best results per column are in \textbf{bold}.}
\label{table:CombinedMetrics}
\end{table*}

\begin{table*}[ht]
\centering
\small
\renewcommand{\arraystretch}{0.9}
\begin{tabular}{l|c|c|c|c|c|c|c|c|c|c|c|c| c}
\hline
& & \multicolumn{6}{c|}{\textbf{OVS-1}} & \multicolumn{6}{c}{\textbf{OVS-2}} \\ 
\hline
\textbf{Dataset} & \textbf{SSIM} & \multicolumn{3}{c|}{\underline{\textbf{MMPMR}}} & \multicolumn{3}{c|}{\underline{\textbf{FMMPMR}}} & \multicolumn{3}{c|}{\underline{\textbf{MMPMR}}} & \multicolumn{3}{c}{\underline{\textbf{FMMPMR}}} \\ 
& & \textbf{0.01\%} & \textbf{0.1\%} & \textbf{1\%} & \textbf{0.01\%} & \textbf{0.1\%} & \textbf{1\%} & \textbf{0.01\%} & \textbf{0.1\%} & \textbf{1\%} & \textbf{0.01\%} & \textbf{0.1\%} & \textbf{1\%} \\ \hline
UFPR & 0.75 & 47.74 & 90.49 & 99.86 & 9.09 & 51,18 & 94.78 & 51.53 & 84.58 & 99.21 & 14.48 & 47.05 & 88.32 \\ 
UBIPr & 0.74 & 10.41 & 74.97 & 99.95 & 0.37 & 28.90 & 96.32 & 30.19 & 74.92 & 99.74 & 5.70 & 36.77 & 91.69 \\ 
MICHE & 0.71 & 32.04 & 63.27 & 99.39 & 4.18 & 15.31 & 88.27 & 23.81 & 54.25 & 95.22 & 2.44 & 15.53 & 69.49 \\ \hline
\end{tabular}
\caption{Cross-dataset evaluation of DOOMGAN's impact on OVS-I and OVS-II at various operating thresholds.}
\label{table:CrossDatasetEvaluation}
\end{table*}

\subsection{Dynamic Weight Adjustment Mechanism}
We employ a dynamic weight adjustment mechanism to balance the contribution of each loss component as follows:

\begin{enumerate}
    \item \textbf{Inverse Loss Calculation}: For each loss $\mathcal{L}_i$, we calculate its inverse:
        \begin{equation}
            \mathcal{L}_i^{-1} = \frac{1}{\mathcal{L}_i + \epsilon}
        \end{equation}
        where $\epsilon$ is a small constant to prevent division by zero.

    \item \textbf{Target Weight Calculation}: We compute target weights based on the inverse losses:
        \begin{equation}
            w_i^{\text{target}} = \frac{\mathcal{L}_i^{-1}}{\sum_j \mathcal{L}_j^{-1}}
        \end{equation}

    \item \textbf{Weight Update}: We adjust the current weights towards the target weights:
        \begin{equation}
            w_i^{\text{new}} = w_i^{\text{current}} + r(w_i^{\text{target}} - w_i^{\text{current}})
        \end{equation}
        where $r$ is the adjustment rate, empirically set to $0.05$ for stable convergence.

    \item \textbf{Weight Normalization}: Finally, we normalize the weights to ensure that they sum up to $1$:
    \begin{equation}
            w_i^{\text{final}} = \frac{w_i^{\text{new}}}{\sum_j w_j^{\text{new}}}
        \end{equation}
\end{enumerate}

\vspace{-0.5em}

This dynamic adjustment allows the model to adaptively focus on the most relevant loss components throughout the training process, leading to more stable and effective optimization. Readers are referred to Section $4$ in the supplementary material for further~\textit{details on our proposed DOOMGAN architecture, the training algorithm, and the chosen hyperparameters}.

\section{Experimental Validations}

\noindent \textbf{Datasets.} We performed evaluations using four established benchmark datasets for visible-light ocular-based analysis tasks: VISOB~\cite{rattani2016icip}, UFPR~\cite{zanlorensi2022new}, UBIPr~\cite{padole2012periocular}, and MICHE~\cite{de2015mobile}. Each dataset presents unique characteristics and challenges. For further details on the datasets, readers are referred to Section $5$ in the supplementary material.


\vspace{0.5em}

\noindent\textbf{Implementation Details and Baselines.} Our proposed DOOMGAN model was trained on ocular images of size $256 \times 256$ and normalized to the range [-1, 1] using the VISOB training dataset. Data augmentation techniques, including random horizontal flipping, center cropping, resizing, and normalization, were applied to enhance dataset diversity and model generalizability. The model was trained using AdamW optimizer ($\beta{_1}$ set to $0.5$) with a learning rate of $2e-4$ for both encoder and generator and $1e-5$ for discriminator and using the batch size of $64$. To avoid overfitting and improve generalization, a weight decay of $1e-5$ was applied to all optimizers, and the learning rate was adjusted using an exponential decay scheduler with a gamma of $0.9998$. To ensure stable training and prevent mode collapse, we used a gradient penalty with a weight of $10$ while computing the discriminator loss. The loss functions used ($\mathcal{L}_{\text{adv}}$, $\mathcal{L}_{\text{MS-SSIM}}$, $\mathcal{L}_{\text{perc}}$, $\mathcal{L}_{\text{rec}}$, $\mathcal{L}_{\text{Id}}$, $\mathcal{L}_{\text{id\_diff}}$) were dynamically weighted during training to effectively balance their contributions, and their final weights obtained after training were $0.0964$, $0.0967$, $0.0965$, $0.5177$, $0.097$, and $0.097$, respectively. $\mathcal{L}_{\text{rec}}$ obtained the highest weightage, as it directly measures the fidelity of the generated image to the original input, ensuring that the morphing process retains essential details and the ocular structure of the corresponding subjects. Pre-trained ocular verification (OVS) models were used to access the vulnerability of generated morphs based on the ResNet-50 backbone using the Arcface loss function~\cite{reddy2020generalizable, reddy2019robust} (\textbf{OVS-I}) and Vision Transformer (\textbf{ViT}) (\textbf{OVS-II}), following studies in~\cite{dosovitskiy2020image, liu2021swin}. The \textbf{Baselines} include traditional image-level ocular morphing techniques, specifically landmark-based methods employing the Delaunay triangulation scheme~\cite{sharma2021image} (using 38 (LM) obtained via $L_g$), alongside GAN-based face morphing models i.e., MorGAN~\cite{damer2018morgan}, StyleGAN~\cite{venkatesh2020can}, VAE-GAN~\cite{larsen2016autoencoding}, MIPGAN~\cite{zhang2021mipgan}(discussed in Section~\ref{related}) are used as baselines for cross-comparison with DOOMGAN. Readers are referred to Section $3$ in the supplementary material for further details on landmark-based ocular image morphing. For MAD defenses, we employ LPQ-SVM~\cite{ahonen2008recognition}, HOG-RF~\cite{dalal2005histograms}, BSIF-SVM~\cite{kannala2012bsif}, and DenseNet~\cite{zhu2017densenet} as representative techniques. These MAD models were selected to encompass both traditional, handcrafted feature-based approaches (LPQ-SVM, HOG-RF, BSIF-SVM) and a deep learning-based method (DenseNet), providing a broad assessment of current morphing detection capabilities against ocular morphing attacks generated using DOOMGAN. Further, all the GAN models and MAD defenses were trained on the VISOB dataset.

\vspace{0.5em}

\noindent\textbf{Standard and Proposed Evaluation Metrics.} We use several metrics in our quantitative experiments. The \textit{Structural Similarity Index Measure} (SSIM), measures the structural similarity between two images with values ranging from $-1$ to $1$ and measures the quality of the generated morphed images. For vulnerability analysis of the morphs to biometric systems, we compute standard metrics, ~\textit{Mated Morphed Presentation Match Rate (MMPMR)} and \textit{Fully Mated Morphed Presentation Match Rate (FMMPMR)}~\cite{scherhag2017biometric,sharma2021image}. MMPMR is the ratio of the number of successful morph attacks to the total number of morph attacks. FMMPMR~\cite{venkatesh2020influence} measures that all contributing subjects must exceed the verification threshold of ocular verification. 

Next, we discuss the \textbf{proposed} IR and Gaze direction metrics explained as follows: \textit{Iris Irregularity~(IR)} quantifies the regularity of the iris shape of the generated ocular morphs using a U-Net-based model for iris boundary segmentation and fitting the ellipse on the segmented boundary using the least square method. Finally, the Boundary Intersection over Union~(BIoU) between the segmented boundary and fitted ellipse represents the iris regularity measure. The segmentation masks from the UBIPr dataset were used to train U-Net for the segmentation of the iris boundary for the estimation of IR metrics. 
\textit{Gaze Direction~(Gaze):} The proposed Gaze metric measures the geometric consistency of the gaze direction in the morphed image. It ensures the generated iris position lies plausibly between the iris coordinates of the two contributing subjects, reflecting a coherent blend rather than an anatomically unrealistic shift. The spatial coordinates of the two irises are obtained using our $L_g$, and evaluated using Euclidean distance. Morph attack Detection Algorithms are evaluated using \textit{Attack Presentation Classification Error Rate~(APCER)} and \textit{Bona Fide Presentation Classification Error Rate~(BPCER)}, which is the proportion of morphed presentations incorrectly classified as bonafide, and vice-versa, respectively. We also report the inference time (Inf.), measured in milliseconds (ms), to generate each morphed image using different models.

\begin{table}[!ht]
    \centering
    \small
    \renewcommand{\arraystretch}{0.9}
    \begin{tabular}{>{\centering\arraybackslash}p{2.1cm}|>{\centering\arraybackslash}p{1.2cm}|>{\centering\arraybackslash}p{1cm}|>{\centering\arraybackslash}p{1.2cm}|>{\centering\arraybackslash}p{1cm}}
        \hline
        \multicolumn{1}{c|}{\textbf{\large{Models}}} & \multicolumn{2}{c|}{\textbf{VISOB}} & \multicolumn{2}{c}{\textbf{UFPR}} \\ \cline{2-5}
        & \textbf{BPCER} & \textbf{D-EER} & \textbf{BPCER} & \textbf{D-EER} \\ \hline
        LPQ-SVM \cite{ahonen2008recognition} & 72.81\% & 26.23\% & 96.02\% & 40.67\% \\ 
        HOG-RF \cite{dalal2005histograms} & 40.35\% & 16.39\% & 99.49\% & 55.96\% \\ 
        BSIF-SVM \cite{kannala2012bsif} & 74.56\% & 22.13\% & 91.03\% & 41.69\% \\ 
        DenseNet \cite{zhu2017densenet} & 36.84\% & 12.30\% & 87.77\% & 39.96\% \\ \hline
    \end{tabular}
    \caption{Performance of various MAD techniques on VISOB and UFPR (cross-dataset) for BPCER @ 5\%APCER}
    \label{tab:performance_metrics}
\end{table}

\begin{table*}[ht]
\centering
\small
\setlength{\tabcolsep}{5pt}
\renewcommand{\arraystretch}{0.85}
\begin{tabular}{l|ccccccc|ccc|cc}
\textbf{Config} & \textbf{Adv.} & \textbf{Perc.} & \textbf{Recon.} & \textbf{MS-SSIM} & \textbf{Identity} & \textbf{Id-Diff} & \textbf{LM} & \textbf{IR$\uparrow$} & \textbf{Gaze$\uparrow$} & \textbf{SSIM$\uparrow$} & \textbf{MMPMR$\uparrow$} & \textbf{FMMPMR$\uparrow$} \\
\hline
Model-1   & \cmark & \cmark & \cmark & \xmark & \xmark & \xmark & \xmark & 0.71 & 0.65 & 0.56 & 82.25 & 40.16 \\
Model-2   & \cmark & \cmark & \cmark & \cmark & \xmark & \xmark & \xmark & 0.86 & 0.67 & 0.64 & 82.91 & 42.91 \\
Model-3   & \cmark & \cmark & \cmark & \xmark & \xmark & \xmark & \cmark & 0.90 & 0.90 & 0.52 & 82.37 & 52.88 \\
Model-4   & \cmark & \cmark & \cmark & \xmark & \cmark & \cmark & \xmark & 0.89 & 0.89 & 0.61 & 88.43 & 63.34 \\
\textbf{DOOMGAN} & \cmark & \cmark & \cmark & \cmark & \cmark & \cmark & \cmark & 0.93 & 0.98 & 0.67 & 99.85 & 96.33 \\
\bottomrule
\end{tabular}
\caption{Ablation study highlighting the impact of each loss function on plausibility, fidelity, quality, and biometric performance at 1\% FMR of the generated morphs evaluated on the VISOB dataset.}
\label{tab:model_comparison}
\end{table*}

\noindent\textbf{Results.} Next, we discuss the results of the morphing model on the VISOB test set and cross-dataset evaluation on UFPR, UBIPr, and MICHE. 


\noindent 1.~\textit{Evaluation on VISOB:} Table~\ref{table:CombinedMetrics} presents a comprehensive comparison of our proposed GAN-based \textbf{DOOMGAN} against both landmark-based (LM) and existing GAN-based morphing models. LM~\cite{sharma2021image} achieved an average SSIM of $0.72$, but exhibits inconsistent structural quality—$0.93$ on one subject versus $0.47$ on the other—due to landmark localization artifacts that introduce irregular iris boundaries, resulting in a poor IR score of $0.7328$ and low gaze consistency (Gaze=$0.5916$). In contrast, DOOMGAN attains SSIM of $0.67$ with an excellent IR of $0.9380$, indicating a relative boost of over 25\% in capturing anatomically regular iris shapes and over 40\% in gaze consistency compared to the LM method.

Further comparison with GAN-based morphing models shows that DOOMGAN consistently outperforms state-of-the-art approaches across all metrics, including SSIM, IR, Gaze, OVS-I and OVS-II (Table~\ref{table:CombinedMetrics}). For OVS-I at 0.01\% FMR, DOOMGAN yields +46.78\% and +42.58\% gains in MMPMR over MIPGAN and MorCode, respectively, and +47.39\% and +42.36\% gains in FMMPMR at 0.1\% FMR. Against conventional GANs (MorGAN, StyleGAN, VAE-GAN), DOOMGAN’s MMPMR improvements exceed 70\% at the strictest thresholds. Under OVS-II, DOOMGAN maintains superiority with more than 50\% MMPMR gains over MIPGAN and MorCode, and over 80\% versus traditional GANs at 0.01\% FMR. Moreover, DOOMGAN outperforms GAN-based morphing methods by approximately 20–46\% in SSIM, 20–58\% in IR, and 29–52\% in gaze consistency. Further, DOOMGAN achieves high fidelity with superior efficiency, by obtaining $2\times$ faster inference time than MorCode, $4\times$ faster than StyleGAN, and over $375\times$ faster than MIPGAN. While legacy methods (MorGAN) obtain faster inference time over DOOMGAN, they generate ineffective, low-quality morphs.

\begin{figure}[ht!]
\centering
\includegraphics[width=0.80\linewidth]{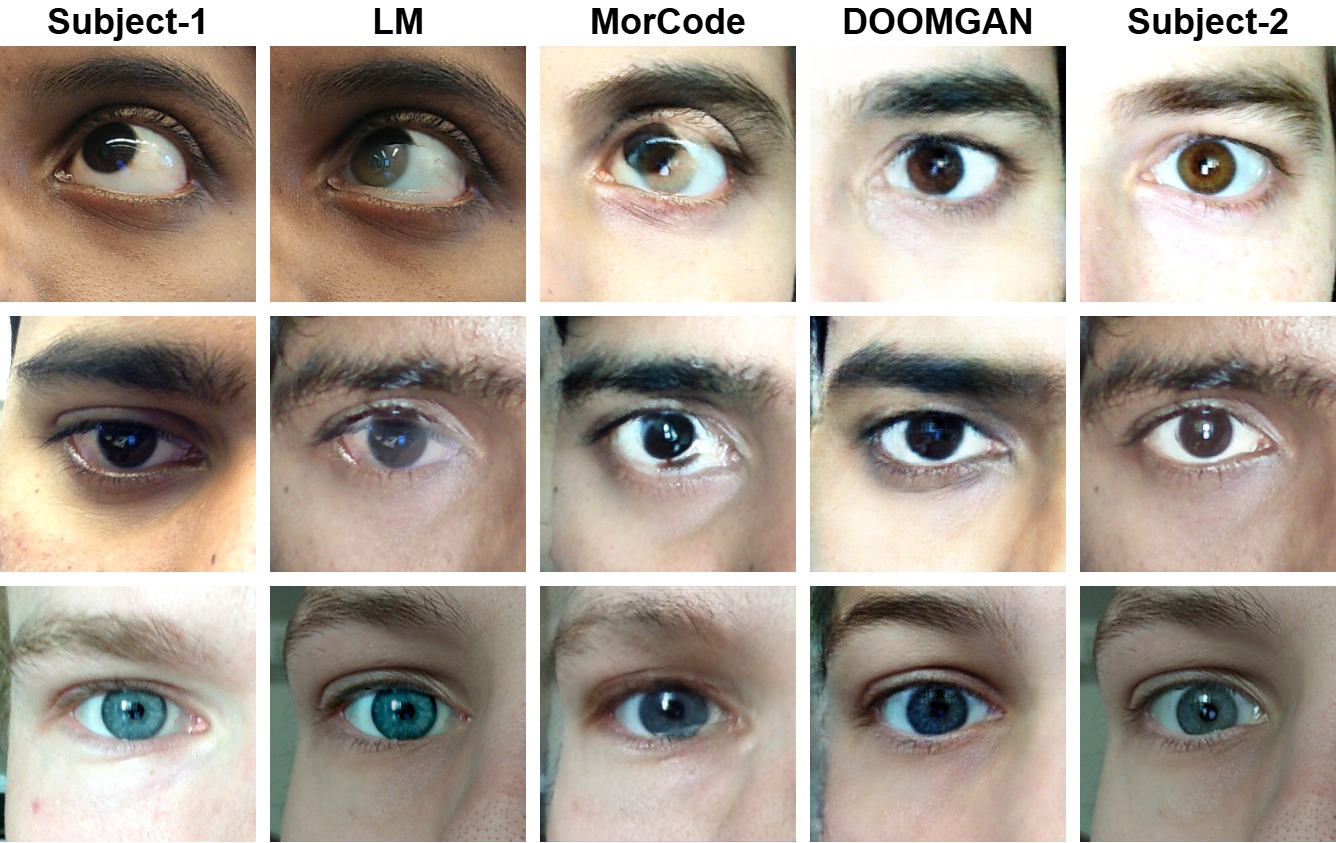}
\captionof{figure}{Comparison of the Landmark-based and best performing GAN-based MorCode model with our proposed DOOMGAN for contributing subject images in an uncontrolled environment, such as off-angle iris and pose, and across demographics.}
\label{fig:compare}
\end{figure}

While conventional landmark-based method (LM) is prone to segmentation artifacts that degrade SSIM, IR, and Gaze, and conventional GANs struggle to capture fine-grained ocular details for reliable OVS evasion, \textbf{DOOMGAN} delivers superior performance across all metrics—demonstrating its strength in generating high-threat morphs through a specialized architecture that preserves intricate ocular biometric features. Figure~\ref{fig:compare} demonstrates the qualitative superiority of the images generated by DOOMGAN. Further, the readers are referred to Section $6$ of the supplementary material for the analysis of OVS similarity scores for both the contributing subjects and the SSIM distributions.

\noindent2.~\textit{Cross-dataset Evaluation on UFPR, UBIPr and MICHE:} Table~\ref{table:CrossDatasetEvaluation} demonstrates DOOMGAN's cross-dataset generalization capabilities across three independent datasets. For OVS-I, UFPR achieves the strongest performance with SSIM of 0.75 and MMPMR of 90.49\% at 0.1\% FMR, while UBIPr and MICHE show varying performance levels (74.97\% and 63.27\%, respectively) due to their distinct image characteristics. Similar trends are observed for OVS-II, with UFPR maintaining robust performance (84.58\% MMPMR at 0.1\% FMR), followed by UBIPr (74.92\%) and MICHE (54.25\%). While FMMPMR scores show expected degradation in cross-dataset scenarios, they maintain reasonable performance, particularly for UFPR (51.18\% at 0.1\% FMR). Notably, all three datasets demonstrate exceptional MMPMR values exceeding 95\% at 1\% FMR, \textit{validating DOOMGAN's cross-dataset attack capability}. The consistently high SSIM scores ($>$0.70) over the baseline VISOB dataset in Table~\ref{table:CombinedMetrics} can be attributed to limited ethnic diversity in the sample populations. Note that while other GAN-based morphing approaches were \textit{evaluated across datasets}, they failed to achieve reasonable FMMPMR scores at more stringent thresholds, and thus are not included in this detailed comparison. For instance, on evaluating StyleGAN morphs on UFPR, UBIPr, and MICHE, the attack success rates at 0.01\% FMMPMR were 0.1\%, 0.8\%, and 0\%, respectively, on OVS-I, and 0.6\%, 1.2\%, and 0\% on OVS-II. For MIPGAN morphs, the corresponding 0.01\% FMMPMR success rates were 0.1\%, 1.3\%, and 0\% on OVS-I, and 1.2\%, 3.2\%, and 0.6\% on OVS-II. DOOMGAN, in contrast, exhibits strong cross-dataset generalization, successfully generating high-threat morphs even on datasets different from its training data.


\noindent3. \textit{Efficacy of Existing MAD:} Table~\ref{tab:performance_metrics} evaluates four prominent MAD techniques on morphed images generated by DOOMGAN, revealing significant vulnerabilities in current detection methods. Traditional feature extractors like LPQ-SVM and BSIF-SVM show poor detection capability even on intra-dataset evaluation on VISOB, with high BPCERs of 72.81\% and 74.56\%, respectively.  The deep learning architecture, DenseNet, obtains moderate performance on VISOB (BPCER: 36.84\%, D-EER: 12.30\%) but fails to generalize well on cross-dataset validation of UFPR (BPCER: 87.77\%).  Results similar to UFPR were obtained for the UBIPr and MICHE datasets as well, but are not included for the sake of space. These results highlight a \textbf{critical gap in current MAD abilities} to detect the generated morphs, particularly in cross-dataset scenarios, emphasizing the need for developing novel ocular-specific MAD techniques, capable of exploiting unique biomarkers inherent to the eye region, such as fine-grained vasculature patterns and periocular texture details, which existing methods may overlook.

\section{Ablation Study}

\begin{figure}[!ht]
    \centering
    \includegraphics[width=0.85\linewidth]{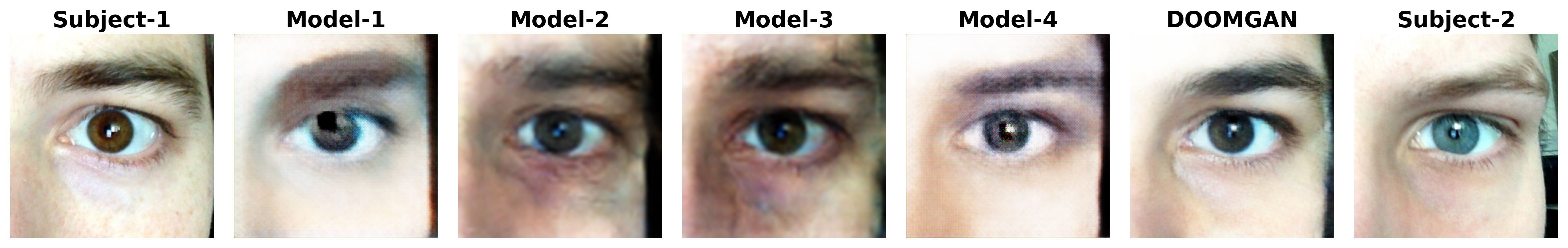}
    \caption{Qualitative Ablation Study across Models.}
    \label{fig:ablate}
\end{figure}

\noindent Table~\ref{tab:model_comparison} evaluates the effectiveness of DOOMGAN's components through systematic isolation testing, maintaining the fundamental trinity of adversarial, perceptual, and reconstruction losses across all configurations. \textit{Model-1}, implementing only these fundamental losses, achieves modest performance (IR 0.71, FMMPMR 40.16\%), reflecting the \textit{lack of explicit structural, anatomical, or identity guidance}. \textit{Model-2} incorporates the MS-SSIM loss, yielding improved structural similarity (SSIM +12.5\% compared to Model-1's 0.56 SSIM), directly attributable to MS-SSIM \textit{optimizing multi-scale structural fidelity}. In contrast, \textit{Model-3} evaluates adding landmark guidance (LM) instead, resulting in significantly improved anatomical plausibility (IR and Gaze reaching 0.90), demonstrating the effectiveness of landmarks in \textit{enforcing geometric consistency}. To assess the contribution of Identity and Identity Difference losses, \textit{Model-4} notably enhances biometric performance (MMPMR 88.43\%, FMMPMR 63.34\%), as these losses directly target the \textit{preservation of identity-related features} crucial for matching. \textit{DOOMGAN}, synergistically combining all components,   demonstrates substantial improvements over individual component configurations. DOOMGAN model achieves a 31\%  improvement in IR (0.93), a 51\% enhancement in Gaze score (0.98), and a 20\% increase in SSIM (0.67). Most notably, it shows remarkable gains in biometric vulnerability metrics, with a 21.4\% improvement in MMPMR (99.85\% ) and an impressive 140\% increase in FMMPMR (96.33\%) compared to the baseline Model-1. These significant performance boosts, as visually demonstrated in Figure~\ref{fig:ablate}, validate the effectiveness of our comprehensive approach in generating high-fidelity ocular morphs while maintaining anatomical consistency and identity preservation.

\vspace{-0.5em}
\section{Conclusion and Future Research Directions}
The threat of morphing attacks can undermine the trustworthiness of the biometric systems. We introduce DOOMGAN, a novel architecture for generating visible-spectrum ocular morphs, achieving state-of-the-art performance through attention- and landmark-guided synthesis with dynamic loss weighting. Extensive evaluations on benchmark datasets and verification systems demonstrate that DOOMGAN-generated morphs significantly compromise system integrity, while existing MAD models remain largely ineffective, exposing a serious security gap. To advance trusted ocular biometrics, future work should focus on: 1) ocular-specific morph generation and detection leveraging unique biomarkers (e.g., vasculature, periocular texture); 2) integrating emerging generative models for broader vulnerability analysis; and 3) enhancing demographic diversity in morphs using advanced techniques like diffusion models.



{\small
\bibliographystyle{ieee}
\bibliography{egbib}
}

\clearpage
\setcounter{page}{1}

\setcounter{figure}{0}
\setcounter{equation}{0}
\setcounter{section}{0} 
\setcounter{table}{0}

\section*{Supplementary: DOOMGAN: High-Fidelity Dynamic Identity Obfuscation Ocular Generative Morphing} 
\section{Existing Methods on Iris Morphing at NIR spectrum} 
Table~\ref{tab:morphing_methods} presents existing studies on iris morphing at the NIR spectrum. 
As can be seen, all the existing studies on iris morphing are conducted in the near-infrared spectrum. At the NIR spectrum, the images are acquired in an illumination-controlled and constrained environment. In contrast, visible light ocular images are acquired in an uncontrolled environment using smartphones. Thus the acquired ocular images represent varying covariates such as off-angle iris, specular reflection, and pose variation. To the best of our knowledge, ocular morphing has never been studied in the visible spectrum.

\begin{table*}[!h]
\centering
\renewcommand{\arraystretch}{1.3}
\begin{tabular}{|p{2cm}|p{2cm}|p{6.5cm}|p{3cm}|}
\hline
\textbf{Reference} & \textbf{Domain} & \textbf{Contribution} & \textbf{Results} \\
\hline
Rathgeb \& Busch \cite{rathgeb2017feasibility} & Iris (Normalized) & Feature-level morphing using stability-based bit substitution for iris-codes to successfully generate morphed iris-codes with maintained biometric utility & \textbf{Attack Success Rate}: 75.27\% to 88.11\% \\
\hline
Erdogan \cite{Erdogan} & NIR Iris (Normalized) & Morphing on normalized iris images using intensity and phase profiles for pixel selection to achieve realistic iris morphs while preserving iris texture patterns & \textbf{Classification Accuracy}: 66.8\% to 76.3\%\\
\hline
Sharma et al. \cite{sharma2021image} & NIR Iris (Unnormalized) & Landmark-based Delaunay triangulation to match the Correspondence to obtain a morph of the two IRIS images & \textbf{SSIM}: 0.31 to 0.49 \textbf{MMPMR}: 17.64 to 97.07\\
\hline
\end{tabular}
\caption{Existing Studies on Iris Image Morphing.}
\label{tab:morphing_methods}
\end{table*}

\section{Proposed Ocular Landmark Generator}
\label{sec:rationale}

\begin{table*}[!ht]
\centering
\renewcommand{\arraystretch}{1.2}
\begin{tabular}{|l|c|c|}
\hline
\textbf{Feature}              & \textbf{Mediapipe IRIS}                          & \textbf{Proposed Landmark Generator}       \\ \hline
\textbf{Input Requirement}    & Requires a complete facial image                      & Works specifically on ocular images        \\ \hline
\textbf{Padding}              & Requires Precise padding                        & Additional padding is not required               \\ \hline
\textbf{Accuracy}             & Less accurate for ocular landmarks               & Highly accurate for ocular landmarks (0.54 MSE)       \\ \hline
\textbf{Ocular Requirement}   & Requires both left and right ocular regions     & Works with any ocular region               \\ \hline
\textbf{Parameterization}     & Challenging to parameterize                     & Easily accessible with helper functions    \\ \hline
\textbf{Landmarks}            & Limited to 10 additional iris landmarks                        & Up to 38 additional ocular-focused landmarks \\ \hline
\textbf{Device Compatibility} & Compatible with CPU and GPU                     & Compatible with CPU and GPU                \\ \hline
\textbf{Model Complexity}     & Pre-trained model with fixed parameters         & Customizable model with flexible parameters \\ \hline
\textbf{Flexibility}          & Limited to predefined landmarks                 & Offers flexible landmark options           \\ \hline
\textbf{Visualization}        & Basic visualization tools                       & Advanced visualization options             \\ \hline
\end{tabular}
\caption{Comparison of Mediapipe IRIS and Proposed Landmark Generator}
\label{table:landmark_comparison}
\end{table*}


The MediaPipe-Iris-based ocular landmark detector requires complete facial images and precise image padding ratios to function effectively, limiting its application in scenarios where only ocular images are available. To address this limitation, we propose the ocular landmark generator, $L_g$, specifically designed to extract the landmarks from the periocular region.


As an initial step, we employed Mediapipe-Iris~\cite{ablavatski2020real} to generate a landmark dataset. While this provided a foundation, most images required manual refinement to ensure precise coordinate placement. From Mediapipe-Iris, we extracted $28$ eye landmarks (arranged in $14$ pairs) and $10$ iris landmarks (in $5$ pairs), which allowed us to accurately define the eye region, iris position, and iris center. The combination of these landmarks yielded $19$ pairs of $(x, y)$ coordinates. To achieve more comprehensive coverage of the periocular region, we interpolated $14$ additional landmark pairs, ultimately creating a robust dataset of $66$ landmarks ($33$ pairs).



Our landmark generator $L_g$ was trained on the VISOB dataset using Mean Squared Error (MSE) loss, with optimization handled by the Adam optimizer and exponential learning rate scheduling to ensure stable convergence. The network outputs 66 neurons, corresponding to our comprehensive landmark set, while maintaining a streamlined architecture focused specifically on ocular features. This specialized design enables high precision and accuracy in landmark detection. Unlike Mediapipe-Iris's fixed landmark configuration, our approach offers enhanced flexibility through the Ocular Landmark Generator, which supports various landmark configurations:

\begin{itemize}
\item \textbf{Center}: Precise iris center coordinates
\item \textbf{Iris}: Dedicated iris-specific landmarks
\item \textbf{Eye}: Detailed eye contour landmarks
\item \textbf{Extension}: Comprehensive landmark set
\item Custom combination of the above options
\end{itemize}

A detailed comparison of our proposed landmark generator and MediaPipe-Iris is presented in Table~\ref{table:landmark_comparison}.


Our solution advances ocular biometric research through three key advantages. First, the enhanced accuracy in landmark detection enables more precise iris and periocular feature extraction, leading to improved recognition rates in biometric systems. Second, unlike existing fixed-point solutions, our flexible landmark configuration allows researchers to optimize landmark placement for specific applications—from dense mapping for detailed morphological analysis to sparse configurations for lightweight deployment. Third, this adaptability enables diverse applications: (a) generating robust embeddings for ocular recognition by providing consistent reference points, (b) facilitating controlled synthetic data generation through precise landmark-guided image manipulation, (c) supporting advanced morphing detection by tracking subtle landmark displacements, and (d) improving ocular region extraction by providing accurate boundary definitions for segmentation tasks.

\section{Landmark-based Ocular Image Morphing}
\label{CV_LM_Morph}


Building on our Ocular Landmark Generator's capabilities, we implemented landmark-based ocular morphing with $38$ points ($19$ pairs) for the LM approach. The landmark-based image morphing process consists of five key steps, established in~\cite{sharma2021image}, discussed as follows:
\begin{enumerate}
\item \textbf{Landmark Extraction (Correspondence)}: Precise landmark extraction from source and target images using our Ocular Landmark Generator, with the number of landmarks serving as a tunable parameter.

\item \textbf{Delaunay Triangulation}: Implementation of Delaunay triangulation on landmark points, creating a comprehensive mesh for proper alignment \cite{chen2004optimal}.

\item \textbf{Triangle Warping}: Computation of affine transformations for each corresponding triangle pair between the images of two contributing subjects.

\item \textbf{Texture Blending}: Smooth interpolation of pixel intensities within the corresponding triangles.

\item \textbf{Seamless Cloning}: Application of advanced blending techniques to achieve natural transitions in the periocular area~\cite{perez2023poisson}.
\end{enumerate}

The mathematical foundation of our morphing process is expressed as follows:
Let $I_s$ and $I_t$ represent the ocular images of the two contributing subjects, respectively, with the morphing parameter $\alpha \in [0, 1]$. For each pixel $(x, y)$ in the morphed image $I_m$:
\begin{equation}
I_m(x, y) = (1 - \alpha) \cdot W_s(I_s, x, y) + \alpha \cdot W_t(I_t, x, y)
\end{equation}

where $W_s$ and $W_t$ denote warping functions mapping pixel coordinates from the morphed images of two contributing subjects based on triangulation and affine transformations.

\begin{figure*}[th]
    \centering
    \includegraphics[scale=0.40]{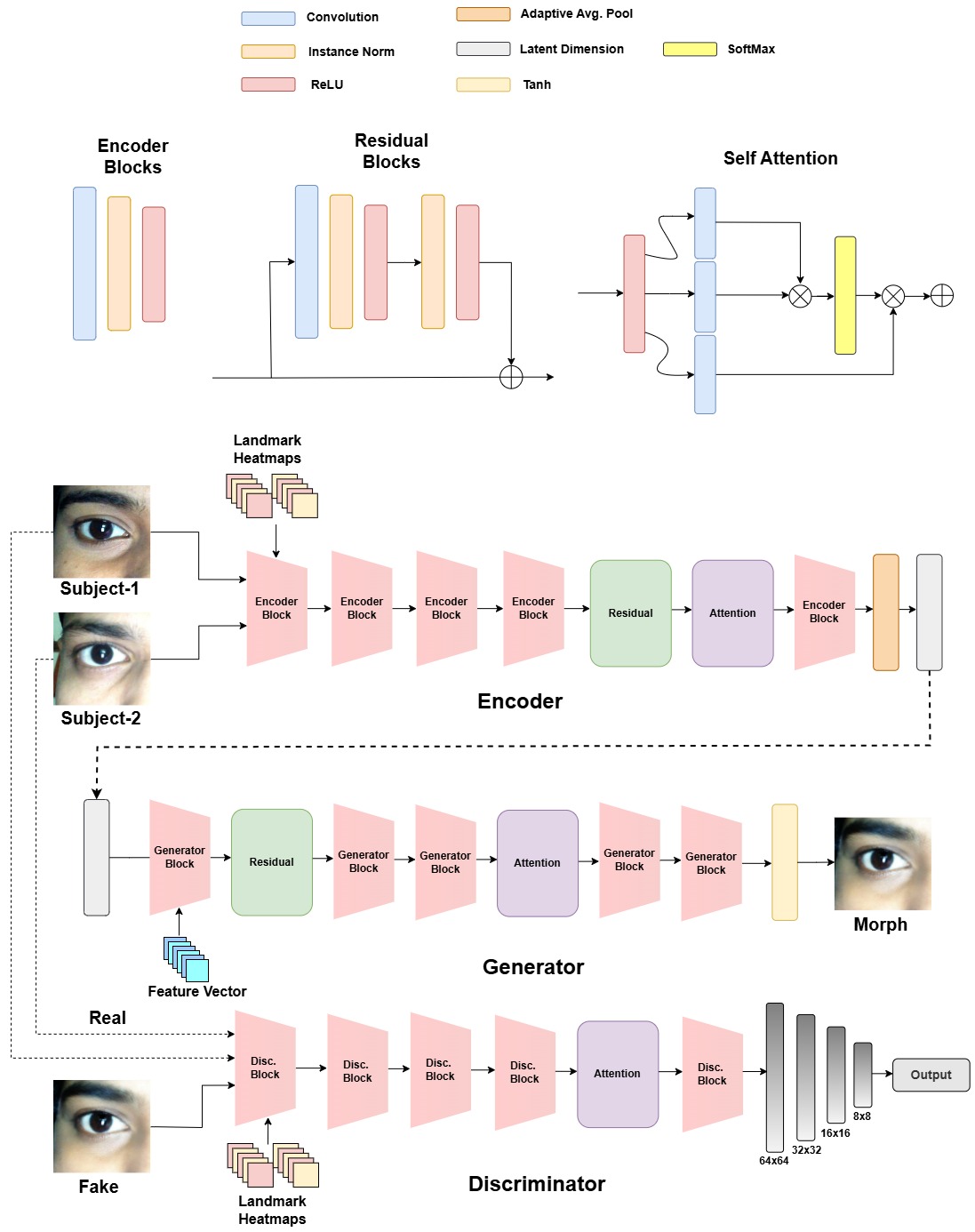}
    \caption{Encoder, Generator, and Discriminator Architecture of the proposed DOOMGAN.}
    \label{fig:OcularMorphGAN}
\end{figure*}

\begin{table*}[htbp]
\centering
\begin{tabular}{l p{5cm} l}
\toprule
\textbf{Component} & \textbf{Input $\rightarrow$ Output Shape} & \textbf{Layer Information} \\ \midrule

\multicolumn{3}{l}{\textbf{Encoder}} \\

\hspace{0.5cm} Conv2d & (\texttt{input + landmarks}) $\rightarrow$ (\texttt{ndf, 64, 64}) & Conv2d: (F=\texttt{ndf}, K=4, S=2, P=1) \\
\hspace{0.5cm} ResidualBlock & (\texttt{ndf * 8, 32, 32}) $\rightarrow$ (\texttt{ndf * 8, 32, 32}) & Residual Block: Conv2d + SN + ReLU \\

\hspace{0.5cm} SelfAttention & (\texttt{ndf * 8, 32, 32}) $\rightarrow$ (\texttt{ndf * 8, 32, 32}) & Self Attention Layer \\
\hspace{0.5cm} AdaptiveAvgPool2d & (\texttt{ndf * 16, 4, 4}) $\rightarrow$ (\texttt{ndf * 16, 1, 1}) & Adaptive Avg Pool \\
\hspace{0.5cm} Conv2d & (\texttt{ndf * 16, 1, 1}) $\rightarrow$ (\texttt{nz, 1, 1}) & Conv2d: (F=\texttt{nz}, K=1, S=1, P=0) \\

\midrule

\multicolumn{3}{l}{\textbf{Generator}} \\

\hspace{0.5cm} ConvTranspose2d & (\texttt{nz + landmark\_features}) $\rightarrow$ (\texttt{ngf * 32, 4, 4}) & ConvTranspose2d: (F=\texttt{ngf * 32}, K=4, S=2, P=1) \\
\hspace{0.5cm} ResidualBlock & (\texttt{ngf * 32, 4, 4}) $\rightarrow$ (\texttt{ngf * 32, 4, 4}) & Residual Block: Conv2d + SN + ReLU \\

\hspace{0.5cm} ConvTranspose2d & (\texttt{ngf * 32, 4, 4}) $\rightarrow$ (\texttt{ngf * 16, 8, 8}) & ConvTranspose2d: (F=\texttt{ngf * 16}, K=4, S=2, P=1) \\
\hspace{0.5cm} SelfAttention & (\texttt{ngf * 16, 8, 8}) $\rightarrow$ (\texttt{ngf * 16, 8, 8}) & Self Attention Layer \\
\hspace{0.5cm} ConvTranspose2d & (\texttt{ngf * 16, 8, 8}) $\rightarrow$ (\texttt{ngf * 8, 16, 16}) & ConvTranspose2d: (F=\texttt{ngf * 8}, K=4, S=2, P=1) \\

\midrule

\multicolumn{3}{l}{\textbf{Discriminator}} \\

\hspace{0.5cm} Conv2d & (\texttt{nc + landmarks}) $\rightarrow$ (\texttt{ndf, 64, 64}) & Conv2d: (F=\texttt{ndf}, K=4, S=2, P=1) \\
\hspace{0.5cm} ResidualBlock & (\texttt{ndf, 64, 64}) $\rightarrow$ (\texttt{ndf, 64, 64}) & Residual Block: Conv2d + SN + ReLU \\

\hspace{0.5cm} Conv2d & (\texttt{ndf, 64, 64}) $\rightarrow$ (\texttt{ndf * 2, 128, 128}) & Conv2d: (F=\texttt{ndf * 2}, K=4, S=2, P=1) \\
\hspace{0.5cm} SelfAttention & (\texttt{ndf * 2, 128, 128}) $\rightarrow$ (\texttt{ndf * 2, 128, 128}) & Self Attention Layer \\
\hspace{0.5cm} Conv2d & (\texttt{ndf * 2, 128, 128}) $\rightarrow$ (\texttt{1, 1, 1}) & Conv2d: (F=1, K=4, S=2, P=0) \\

\bottomrule
\end{tabular}
\caption{Model Architecture for Generator, Discriminator, and Encoder}
\label{tab:model_components}
\end{table*}

\begin{algorithm}
    \caption{DOOMGAN Training with Dynamic Weight Adjustment}
    \label{algorithm:training_loop}
    \begin{algorithmic}[0]
        \State Initialize lists for tracking losses: $Losses_{(DGE)} \gets []$
        \State Initialize $gp_w \gets 10$ \Comment{Gradient Penalty weight}
        \State Initialize $LE, G, D, E$ \Comment{LM Enc., Gen, Disc., Img Enc.}
        \State Initialize optim, and sched for $LE, G, D, E$ 
        \State Initialize $w_{adj}$ 
        
        \Comment{dynamic weight adjuster with initial weights}

        \For{epoch = 1 to $N$}
            \For{each batch in dataloader}
                \State Load $reals$, $lm \gets data$
                \State Generate landmark heatmaps: $LH \gets heatmap(lm)$
                \State Extract landmark features: $LF \gets LE(lm)$
                
                \State \textit{\textbf{Update Discriminator:}}
                \State $errD_{real} \gets \text{D}(reals, LH)$ 
                
                \State $noise \gets E(reals, LH)$
                
                \State $fake_{img} \gets G(noise, LF)$
                
                \State $errD_{fake} \gets \text{D}(fakes, LH)$
                
                \State $gp \gets \text{GP}(reals, fakes, LH)$
                
                \State $errD \gets errD_{real} + errD_{fake} + gp_w \cdot gp$ 
                
                \State $errD.backward()$
                \State $optD.step()$

                \State \textit{\textbf{Update Generator and Encoder:}}
                \State Zero grads for $G$, $E$, $LE$
                \State Recompute $fake_{img} \gets G(noise, LF)$
                \State $errG_{adv} \gets - \text{mean}(\text{D}(fakes, LH))$ 
                
                \State $errG_{perc} \gets \text{Perc.}(fakes, reals)$ 
                
                
                \State $errG_{recon} \gets \text{MSE}(fakes, reals)$ 
                

                \State $errG_{ms\_ssim} \gets \text{MS\_SSIM}(fakes, reals)$ 
                
                
                \State $id_{diff} \gets \text{ID-Diff}(reals, fakes)$ 

                \State $id_{loss} \gets \text{ID}(reals, fakes)$
                

                \State \textit{\textbf{Dynamic Weighting, Backpropagation, Logging:}}
                
                \State $w_{adj} \gets w_{adj}.update()$


                
                \State $errG \gets w_{adj}[adv] \cdot errG_{adv} + w_{adj}[Perc]\cdot errG_{Perc} + $
                \State $ \dots + w_{adj}[ID] \cdot errG_{ID}$ 
                
                \Comment{Total Loss with Updated Weights}
                
                \State $errG.backward()$
                \State $optim.step()$ for $G, E, LE$
                

                \If{$epoch \mod 10 = 0$}
                    \State Logging and Check-pointing
                \EndIf
            \EndFor
            \State $Sched.step()$ for $LE, G, D, E$
        \EndFor
    \end{algorithmic}
\end{algorithm}

\section{DOOM-GAN Architecture and Training Algorithm}
\label{Protocol}

Figure~\ref{fig:OcularMorphGAN} shows the architecture of the Encoder and Generator of the DOOMGAN model, and Table \ref{tab:model_components} shows the network flow of each layer. The notations associated with the Table are represented as shown below.

\begin{itemize}
    \item \textbf{ngf:} Size of feature maps in the generator (64)
    \item \textbf{ndf:} Size of feature maps in the discriminator (64)
    \item \textbf{nz:} Size of latent vector (200)
    \item \textbf{nc:} Number of channels in the input images (3)
    \item \textbf{batch\_size:} Batch size during training (64)
    \item \textbf{image\_size:} Spatial size of training images (256)
    \item \textbf{lr\_e, lr\_g, lr\_d:} Learning rates for encoder, generator, and discriminator (0.0002, 0.0002, 0.00001)
    \item \textbf{beta1:} Beta1 hyperparameter for Adam optimizer (0.5)
\end{itemize}

 As shown in Figure~\ref{fig:OcularMorphGAN} and detailed in Table~\ref{tab:model_components}, DOOMGAN's architecture comprises three main components: an Encoder ($E$), a Generator ($G$), and a Discriminator ($D$). The $E$ processes the input ocular image, combined with landmark heatmaps ($h_l$, to create a latent space representation.  It starts with a convolutional layer (Conv2d) that takes the concatenated input and landmarks, outputting a feature map of shape (\texttt{ndf, 64, 64}).  The $E$ then uses a series of operations, including a ResidualBlock (\texttt{ndf * 8, 32, 32} to \texttt{ndf * 8, 32, 32}), and a SelfAttention layer to capture both local and global feature relationships.  An AdaptiveAvgPool2d layer reduces the spatial dimensions to (\texttt{ndf * 16, 1, 1}), and a final Conv2d layer produces the latent vector \texttt{nz} with a shape of (\texttt{nz, 1, 1}).

 The $G$ reconstructs the morphed image from the $E$ latent vector (\texttt{nz}) and landmark features. It begins with a transposed convolutional layer (ConvTranspose2d) that upsamples the input (\texttt{nz + landmark\_features}) to (\texttt{ngf * 32, 4, 4}). A ResidualBlock maintains the feature map shape, followed by another ConvTranspose2d layer, increasing the resolution to (\texttt{ngf * 16, 8, 8}).  A SelfAttention layer is applied, and a final ConvTranspose2d layer outputs a feature map of shape (\texttt{ngf * 8, 16, 16}).  The $D$ distinguishes between real and generated images.  It starts with a Conv2d layer processing the input and landmarks, resulting in (\texttt{ndf, 64, 64}).  A ResidualBlock is then applied, followed by another Conv2d layer that increases the feature channels and reduces spatial dimensions to (\texttt{ndf * 2, 128, 128}). A SelfAttention layer and a final Conv2d layer produce the discriminator's output with a shape of (\texttt{1, 1, 1}), indicating the authenticity of the input.

The training algorithm for our proposed generative network is described in Algorithm \ref{algorithm:training_loop}. We use a fixed gradient penalty weight ($\lambda_{GP}$) set to $10$. The Landmark Encoder, Image Encoder, Generator, and Discriminator are initialized, along with their optimizers and schedulers. In the iterative loop, we load the real images along with their respective landmarks. The landmark heat maps are used to supervise the encoder and discriminator by providing them with additional information. The landmark encoder provides the landmark feature vectors used to guide the generator. The backpropagation involves the continuous back-and-forth training of the generated morphs with the real images to obtain the ideal output while balancing all the losses described previously. Dynamic weighting plays a critical role in adjusting the loss functions appropriately to obtain a suitable equilibrium during training.

\begin{figure*}[!h]
    \centering
    \includegraphics[scale=0.33]{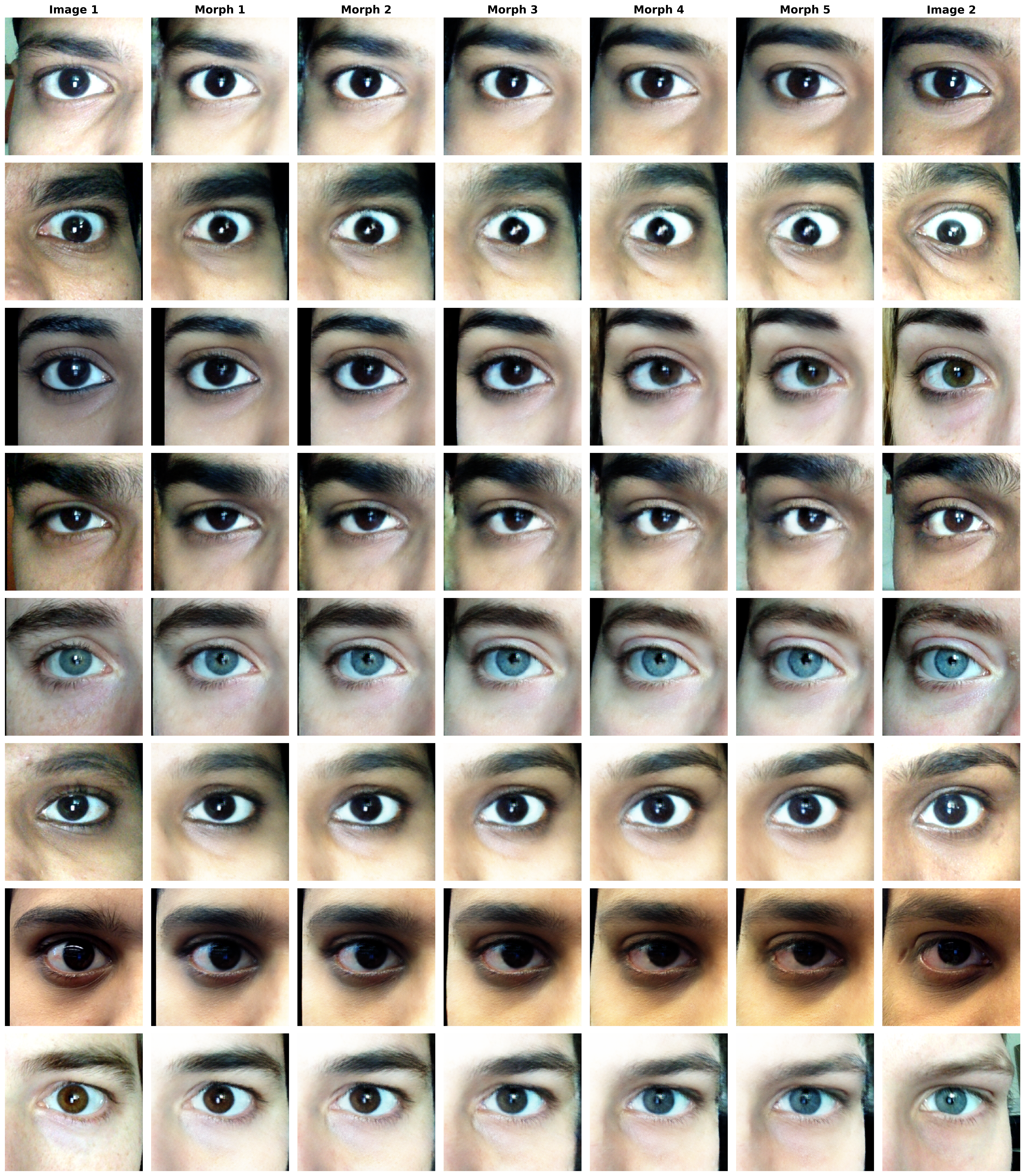}
     \caption{Disentangled Morphs for contributing subjects across different interpolation latent points.}
     \label{fig:Final_Fig}
 \end{figure*}

The proposed generative model was implemented using PyTorch and trained on $2$ NVIDIA A$5000$ $32$GB GPUs. The training process took approximately $5$ days to converge. Figure~\ref{fig:Final_Fig} shows disentangled morphs for contributing subjects across different interpolation points.

\section{Dataset Details}
The VISOB dataset comprises images from $585$ subjects, captured using a variety of mobile phone cameras under diverse lighting conditions. The UFPR dataset features images of 1,122 subjects, acquired across 196 different mobile devices, representing a broad range of acquisition hardware. The UBIPr dataset provides a substantial collection of more than 11,000 images, notable for its inclusion of detailed segmentation masks that delineate key periocular regions (skin, eyebrows, sclera, and iris). The MICHE dataset contributes ocular images from 75 subjects, obtained using various mobile devices in uncontrolled environments. The combination of these datasets ensures a comprehensive and rigorous evaluation of our model's performance across a wide spectrum of conditions.

\section{Analysis of Similarity Score Distributions and Image Quality Metrics}
\begin{figure}[htbp]
  \centering
  \begin{subfigure}[b]{0.5\linewidth}
    \centering
    \includegraphics[width=\linewidth]{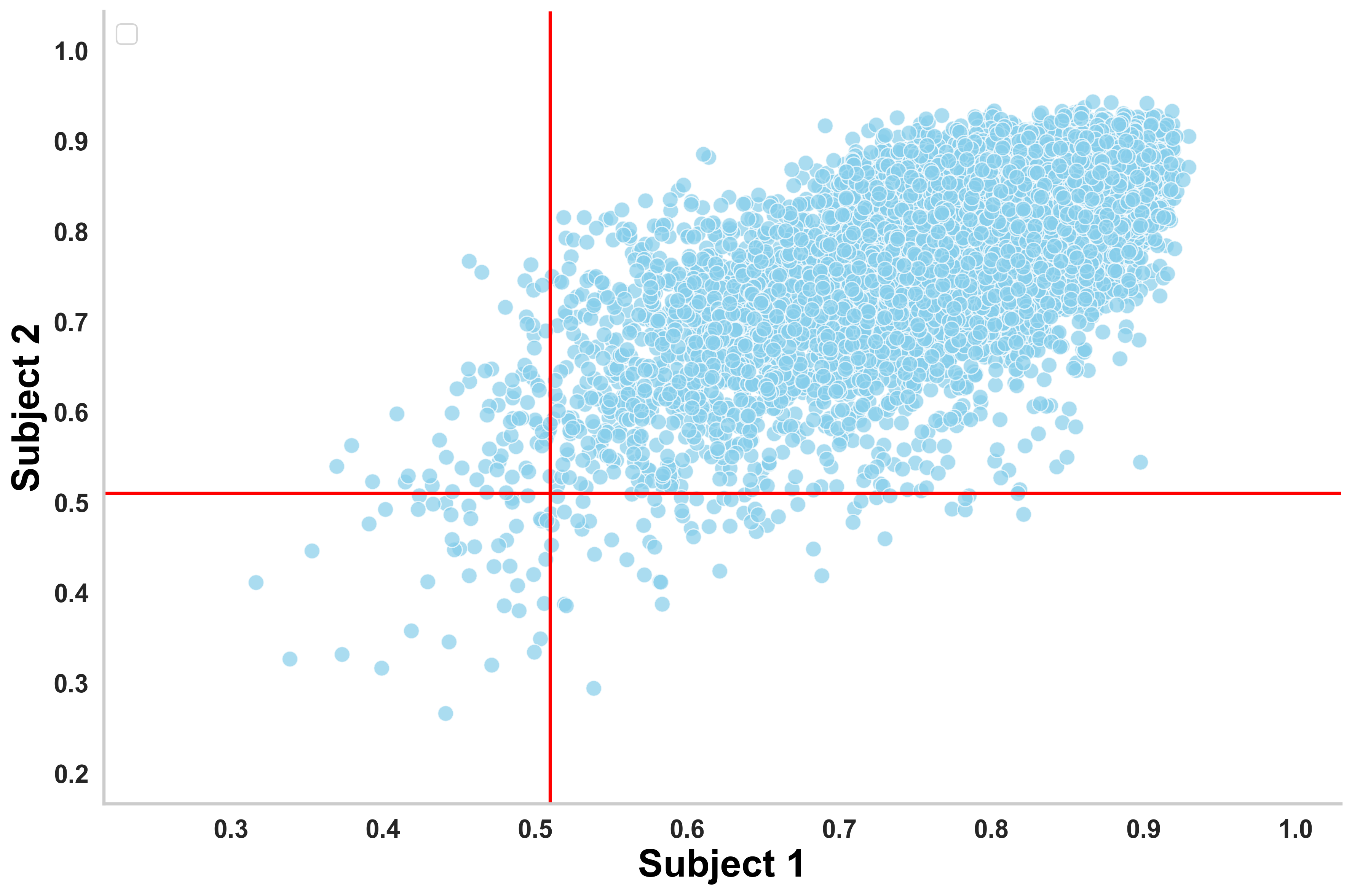}  
    \caption{}
    \label{fig:similarity-all}
  \end{subfigure}%
  \begin{subfigure}[b]{0.5\linewidth}
    \centering
    \includegraphics[width=\linewidth]{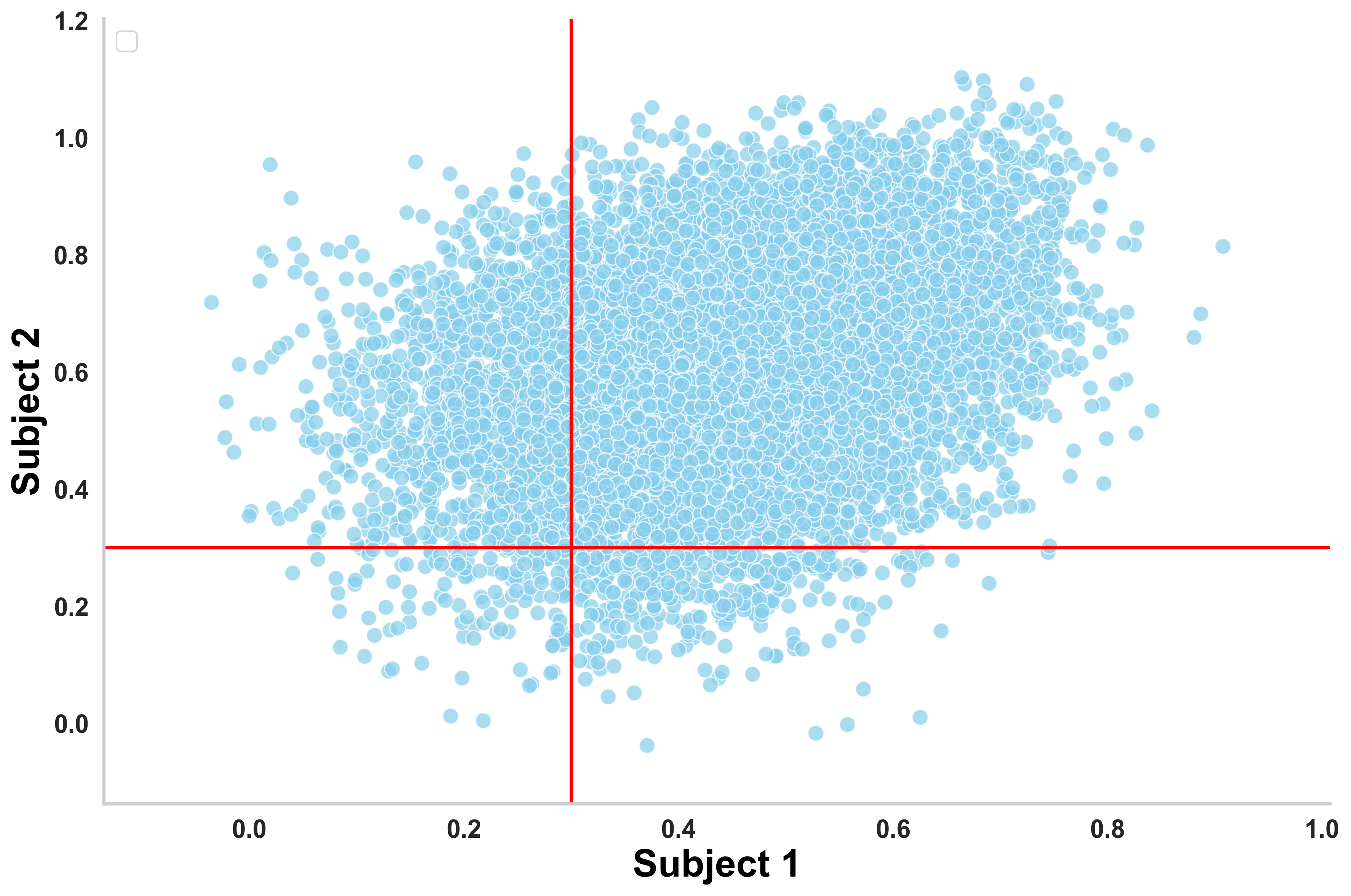}  
    \caption{}
    \label{fig:similarity-perfect}
  \end{subfigure}

  \vskip\baselineskip 

  \begin{subfigure}[b]{0.5\linewidth}
    \centering
    \includegraphics[width=\linewidth]{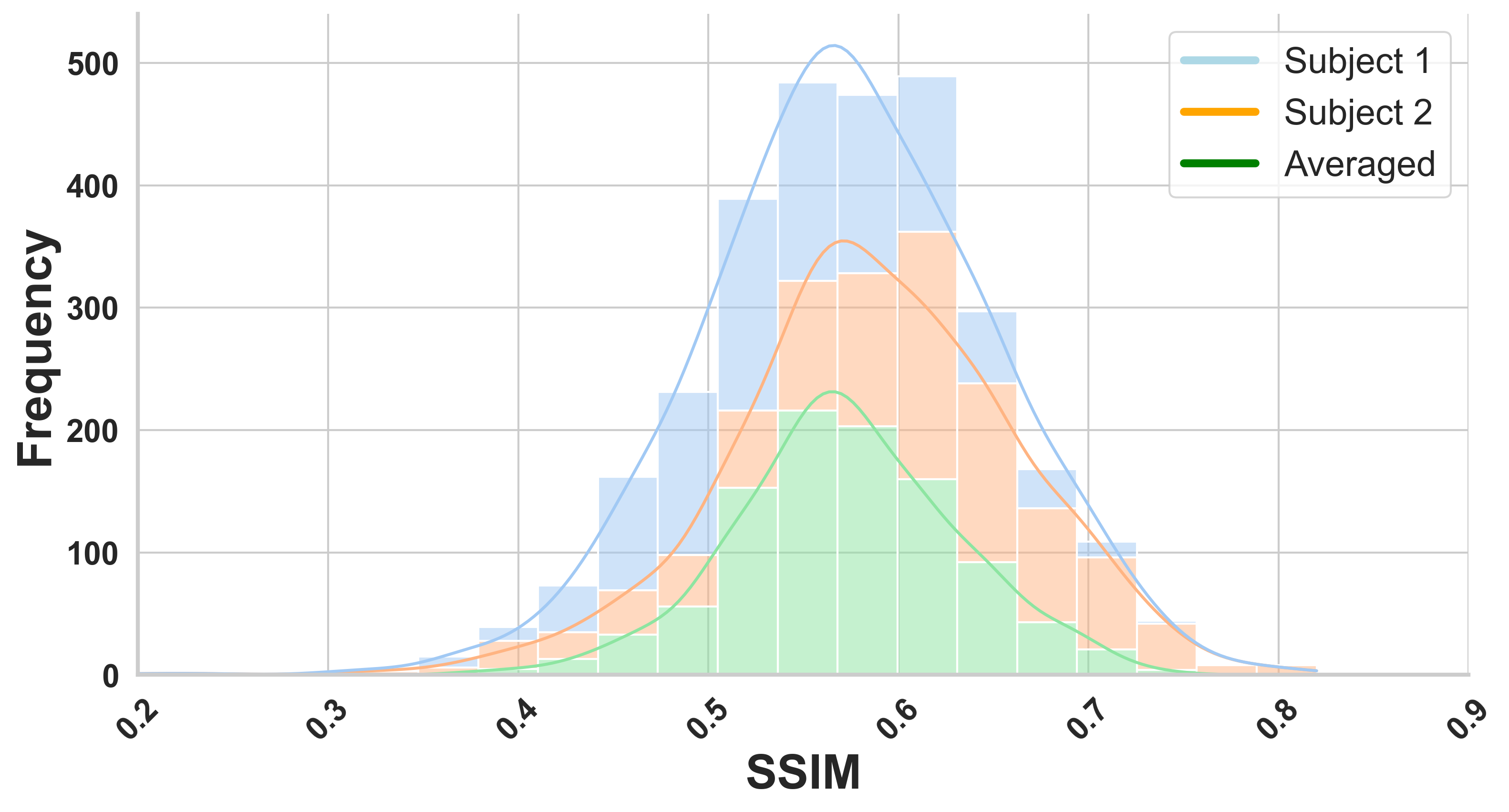}  
    \caption{}
    \label{fig:ssim-visob}
  \end{subfigure}%
  \begin{subfigure}[b]{0.5\linewidth}
    \centering
    \includegraphics[width=\linewidth]{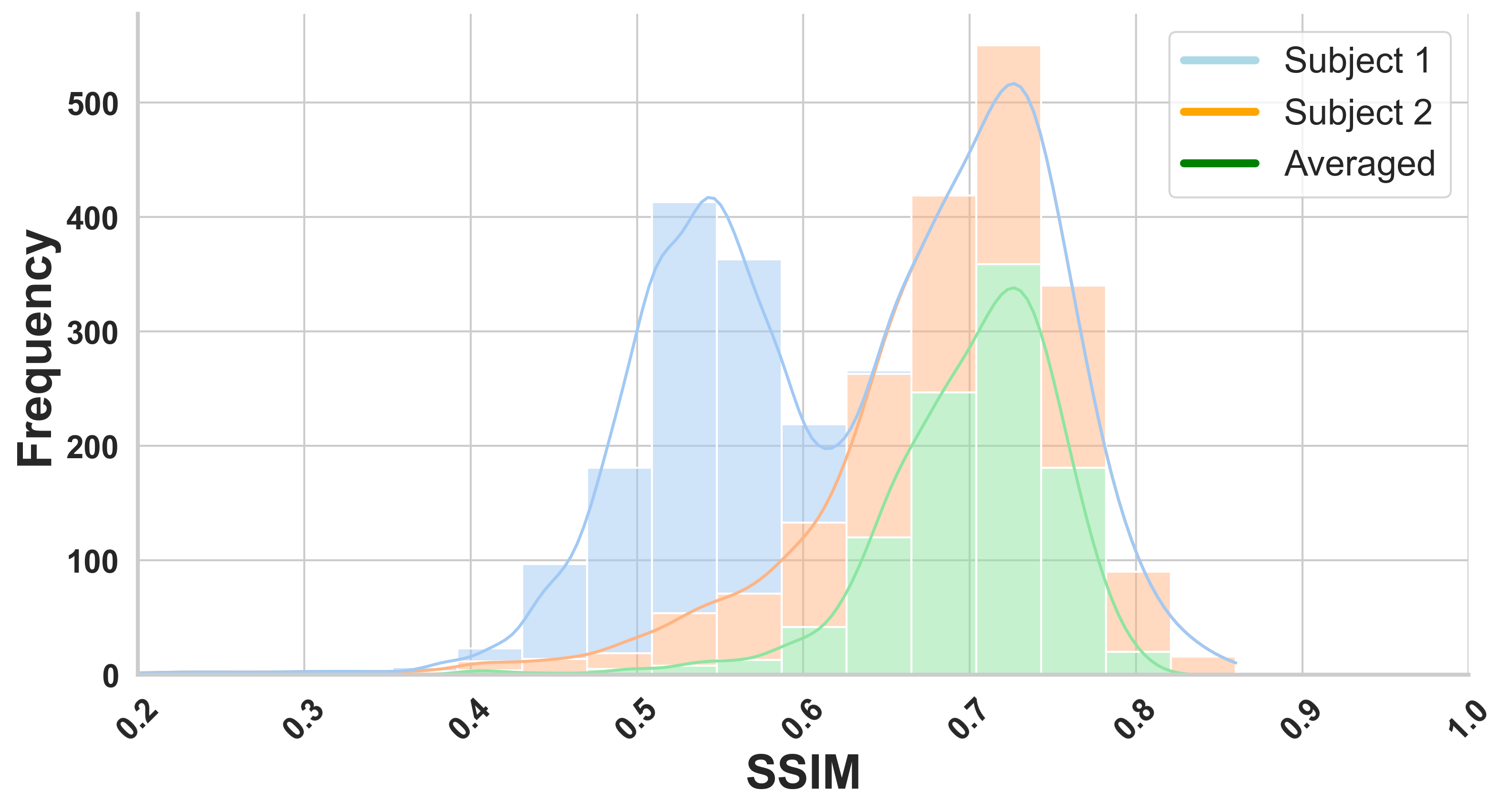}  
    \caption{}
    \label{fig:ssim-ufpr}
  \end{subfigure}
  \caption{Top: Scatter plot of similarity scores on the VISOB dataset for (a) OVS-I and (b) ViT OVS-II. Bottom: Spread of SSIM scores for (c) VISOB and (d) UFPR, for both contributing subjects.}
  \label{fig:comparison}
\end{figure}
~Figure~\ref{fig:comparison} (a) and (c) show the scatter plot and the bar plot of the cosine similarity-based matching scores (using the OVS-I and OVS-II models, respectively) obtained and the SSIM scores of the generated morphed images using our proposed model for the two contributing subjects. The similarity score distributions (Figure \ref{fig:ssim-ufpr}) reveal a skewed pattern where one contributing subject achieves higher scores while the other shows lower performance, contrasting with VISOB's more uniform distribution across both subjects.



\end{document}